\documentclass[runningheads]{llncs}
\usepackage[T1]{fontenc}
\usepackage{graphicx,verbatim}
\usepackage[switch]{lineno}
\usepackage{color,soul}
\usepackage{booktabs}
\usepackage{amssymb}
\usepackage{makecell}
\usepackage{caption}
\begin{document}
%
% \title{\underline{MELON}: \underline{M}ultimodal Mixture-of-\underline{E}xperts for \underline{L}ong-Term M\underline{o}bility Estimatio\underline{n} using in Intensive Care Unit}
\title{MELON: \underline{M}ultimodal Mixture-of-\underline{E}xperts with Spectral-Temporal Fusion for \underline{L}ong-Term M\underline{O}bility Estimatio\underline{N} in Critical Care}
%
\begin{comment}  %% Removed for anonymized MICCAI 2025 submission
\author{First Author\inst{1}\orcidID{0000-1111-2222-3333} \and
Second Author\inst{2,3}\orcidID{1111-2222-3333-4444} \and
Third Author\inst{3}\orcidID{2222--3333-4444-5555}}
%
\authorrunning{F. Author et al.}
% First names are abbreviated in the running head.
% If there are more than two authors, 'et al.' is used.
%
\institute{Princeton University, Princeton NJ 08544, USA \and
Springer Heidelberg, Tiergartenstr. 17, 69121 Heidelberg, Germany
\email{lncs@springer.com}\\
\url{http://www.springer.com/gp/computer-science/lncs} \and
ABC Institute, Rupert-Karls-University Heidelberg, Heidelberg, Germany\\
\email{\{abc,lncs\}@uni-heidelberg.de}}

\end{comment}

\author{Jiaqing Zhang\inst{1} \and
Miguel Contreras\inst{2} \and
Jessica Sena\inst{2} \and
Andrea Davidson \inst{3} \and
Yuanfang Ren\inst{3} \and
Ziyuan Guan\inst{3} \and
Tezcan Ozrazgat-Baslanti\inst{3} \and
Tyler J. Loftus\inst{4}\and
Subhash Nerella\inst{2} \and
Azra Bihorac\inst{3} \and
Parisa Rashidi\inst{2}
}

\authorrunning{J. Zhang et al.}

\institute{Department of Electrical and Computer Engineering, University of Florida, Gainesville, FL,
USA \and
J. Crayton Pruitt Family Department of Biomedical Engineering, University of Florida, Gainesville, FL, USA\and
Department of Medicine, University of Florida, Gainesville, FL, USA\and
Department of Surgery, University of Florida, Gainesville, FL, USA
}

\maketitle              % typeset the header of the contribution
\begin{abstract}
Patient mobility monitoring in intensive care is critical for ensuring timely interventions and improving clinical outcomes. While accelerometry-based sensor data are widely adopted in training artificial intelligence models to estimate patient mobility, existing approaches face two key limitations highlighted in clinical practice: (1) modeling the long-term accelerometer data is challenging due to the high dimensionality, variability, and noise, and (2) the absence of efficient and robust methods for long-term mobility assessment. To overcome these challenges, we introduce MELON, a novel multimodal framework designed to predict 12-hour mobility status in the critical care setting. MELON leverages the power of a dual-branch network architecture, combining the strengths of spectrogram-based visual representations and sequential accelerometer statistical features. MELON effectively captures global and fine-grained mobility patterns by integrating a pre-trained image encoder for rich frequency-domain feature extraction and a Mixture-of-Experts encoder for sequence modeling. We trained and evaluated the MELON model on the multimodal dataset of 126 patients recruited from nine Intensive Care Units at the University of Florida Health Shands Hospital main campus in Gainesville, Florida. Experiments showed that MELON outperforms conventional approaches for 12-hour mobility status estimation with an overall area under the receiver operating characteristic curve (AUROC) of 0.82 (95\%, confidence interval 0.78-0.86). Notably, our experiments also revealed that accelerometer data collected from the wrist provides robust predictive performance compared with data from the ankle, suggesting a single-sensor solution that can reduce patient burden and lower deployment costs. Additionally, we revealed a strong correlation between patient mobility and delirium, which highlight the importance of building a real-time monitoring system.

\keywords{Accelerometer \and Multimodal  \and Spectrogram \and Mobility \and Intensive Care Unit \and Mixture of Expert \and Patient Monitoring.}
% Authors must provide keywords and are not allowed to remove this Keyword section.

\end{abstract}

\section{Introduction}
\label{sec:intro}

Patients in the Intensive Care Unit (ICU) often experience prolonged immobility, increasing the risk of developing ICU-acquired weakness (ICU-AW) and cognitive impairment, such as delirium \cite{chambers2009physical}. Frequent patient mobility assessments are pivotal for facilitating timely interventions and mitigating adverse outcomes \cite{schweickert2007icu,barr2013clinical,tipping2017effects}. Several functional mobility tools have been developed and implemented in clinical practice, such as the ICU Mobility Scale\cite{hodgson2014feasibility} and the Johns Hopkins Highest Level of Mobility Scale \cite{hoyer2018toward}. However, these tools require manual administration by ICU staff, which can be time-consuming and prone to documentation errors. Additionally, manual assessments often lack the granularity needed to capture subtle changes in mobility over time. Autonomous sensor-based approaches offer a promising alternative, enabling continuous, objective, and high-resolution patient mobility monitoring while reducing the workload on healthcare providers.

Accelerometry-based sensors are a commonly used option for capturing physical activity levels, posture changes, and mobility trends over time in a non-invasive manner \cite{schwab2020actigraphy,yang2010review,sena2024wearable}. However, directly modeling raw time-series accelerometer data presents several challenges due to the inherent complexity of the signals. The data consists of long sequences, making it computationally demanding to process and analyze. Additionally, its non-stationary nature—where statistical properties change over time—complicates pattern recognition and model generalization. Despite the development of models that can efficiently handle long sequences of time-series input \cite{liu2025examiningadaptingtimemultilingual,shi2024time,han2024length}, the existing approaches that rely on time-series accelerometry data often fail to capture the nuances of mobility fully. Recent studies have explored converting sequential sensor data into image representations, leveraging the hierarchical feature extraction capabilities of pre-trained vision models \cite{yu2025robust}. For instance, recurrence plots—a method transforming temporal dynamics into 2D texture patterns—have been successfully adopted in works such as \cite{lu2019robust} and \cite{lew2023human}, achieving state-of-the-art performance in activity recognition tasks. Alternatively, frequency-domain representations like spectrograms generated via Short-Time Fourier Transform (STFT) have also proven effective. Studies by \cite{ito2018application} and \cite{sassi2023spectrogram} demonstrate that CNN-based models trained on spectrograms can robustly identify complex activities through learned spectral-temporal patterns.
However, most of these models are developed and validated in general domains, and they do not address the specific needs of the ICU setups, which exhibit different activity patterns \cite{sena2024wearable} and long-term assessment. 

\begin{figure}[t]
\includegraphics[width=\textwidth]{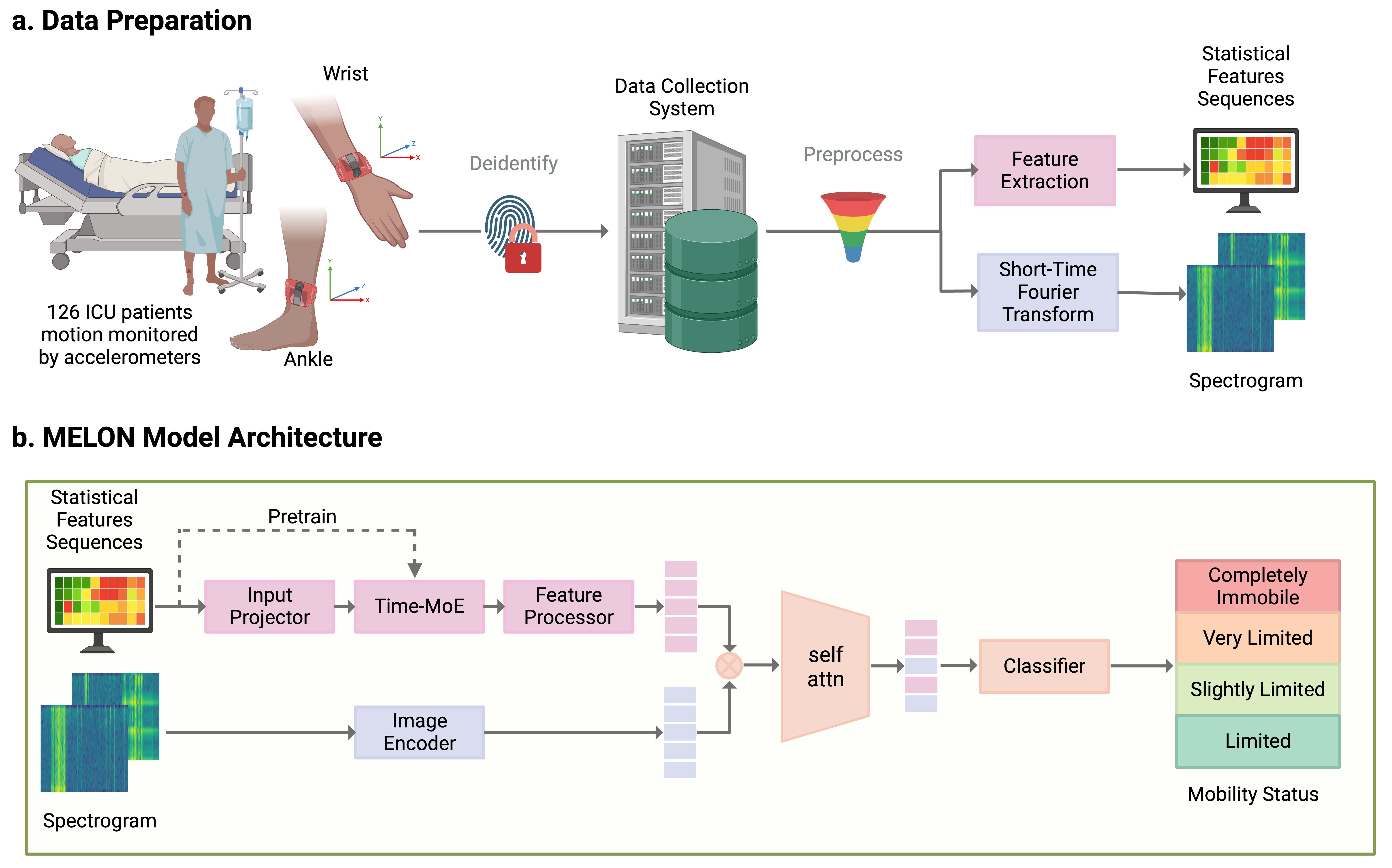}
\caption{In the data preparation stage (a), we collected data from two accelerometers positioned at the patient's nondominant wrist and ankle. Data was then deidentified and stored in a secure server. We generated spectrograms using the Short-Time Fourier Transform and also extracted five features per minute with a 30-second overlap over 12 hours. The MELON model has a dual-branch structure, as shown in (b). The embeddings from the two branches are then fed into a self-attention layer to generate fusion embeddings. Then, the fused embeddings are processed in the classifier for mobility status prediction.} \label{fig1}
\end{figure}

In this study, we propose MELON (Multimodal mixture-of-Experts with spectral-temporal fusion for Long-term mObility estimatioN), a novel dual-branch multimodal architecture for patient mobility assessment Fig \ref{fig1}. The proposed architecture integrates two complementary input modalities: statistical accelerometer features sequence and spectrogram representation images of raw accelerometry data. To effectively process these modalities, we employ a pre-trained mixture-of-experts (Time-MoE) to model statistical feature sequences and an image encoder to extract patterns from frequency domain representations. By combining these modalities, we leverage their complementarity—statistical features offer structured and fine-grained variations, whereas spectrograms enhance the ability to detect global temporal dependency—resulting in a more comprehensive representation of patient mobility. To the best of our knowledge, this is the first study that combines the spectrogram of accelerometer data and sequences of statistical features using a dual-branch multimodal architecture. Our main contributions are:

\begin{description}
\item[(1)] We propose MELON, a novel dual-branch multimodal framework designed to model accelerometer data by processing spectrogram images and their corresponding long-term statistical feature sequences using ResNet and Mixture-of-Experts structure respectively for accurate patients' mobility status prediction in the ICU.

\item[(2)] We performed extensive comparisons using data from 126 ICU patients to highlight the robustness of the proposed method. Our experiments demonstrated that the combination of spectral and temporal information of the accelerometer data could boost the prediction power.

\item[(3)] We accessed multiple analyses to provide practical recommendations in evaluating patients' mobility status.
\end{description}

\section{Methods}
\label{sec:methods}

In this study, we proposed a multimodal dual-branch model, MELON, trained on spectrogram images $I$ and their corresponding sequences of statistical features $A$ extracted from raw accelerometer data to predict patients' mobility measurements over a long-term 12-hour window. For mobility, we had four classes, i.e., completely immobile, very limited, slightly limited, and no limitation. Fig \ref{fig1} illustrates the architecture of our proposed pipeline. 

\subsection{Accelerometer Data Preprocessing}

Given the three-axis raw accelerometer data at 20 Hz for the spectral information, we first transformed the signal into a time-frequency representation using a Short-Time Fourier Transform (STFT). We set the sampling frequency to 20 Hz, a segment length of 64 samples, and an overlap of 32 samples to achieve a balanced resolution in both time and frequency domains. After computing the spectrogram $S$, we apply a logarithmic transformation to compress the dynamic range. The resulting logarithmic spectrogram is then normalized to the $[0,255]$ range and converted to images $I$.

For the temporal information, first, we extracted five features following the prior work \cite{zhang2024mango} which has been proven to be efficient to represent the temporal information: the mean and standard deviation of the vector magnitude, the mean and standard deviation of the angle between the x-axis and the vector magnitude, and the domain frequency. Then, we constructed 12-hour-long feature sequences by (1) calculating these five features for each minute within the 12-hour window, (2) padding missing values when the accelerometer sensor was temporarily removed or the recording time was shorter than 12 hours, and (3) employing a 30-second overlap between adjacent segments. This process produced a feature sequence with 1440 time steps—each corresponding to one minute of data (with 30-second overlaps)—for each of the five features $A\in \mathbb{R} ^{1440\times5}$.

\subsection{Multimodal Dual-Branch Architecture}

We provide the spectral feature images $I$ and the temporal feature sequences $A$ to the MELON model, which consists of three main components: (1) image encoder pre-trained on ImageNet dataset, (2) Time-MoE encoder pre-trained on our accelerometer statistical sequence data, and (3) attention fusion block and classifier.

\subsubsection{Image Encoder}

To encode spectrogram images, we adopt a ResNet model pre-trained on ImageNet as the backbone image encoder. The original fully connected layer is replaced with a tailored projection module. This module consists of a linear layer that projects the high-dimensional feature vector into a 512-dimensional space, followed by a ReLU activation and dropout layer. The embedding of spectrogram is donated as $I_{embed}\in\mathbb{R}^{1\times\hat{D}}$.

\subsubsection{Time-MoE Encoder}

Following Shi et al. \cite{shi2024time}'s work, we utilize an input project $f_{proj}$ and point-wise tokenization for generating embeddings of the accelerometer statistical sequences to ensure the completeness of temporal information. Then a SwiGLU layer is used to process each element of sequences \cite{shazeer2020glu}:

\begin{equation}
    A_{hidden}=SwiGLU(A)=Swish(Wf_{proj}(A))\otimes (Vf_{proj}(A)),
\end{equation}

where $W\in R^{D\times 1}$ and $V\in R^{D\times 1}$ are two learnable parameters, $Swish$ is an activation function introduced by Ramachandran et al. \cite{ramachandran2017searching}, and $D=128$ is the hidden size. Next, the projected $A_{hidden}$ is fed into a MoE transformer block, which consists of a stack of transformer decoder layers augmented with a sparse MoE feed-forward network. In each decoder layer, the self-attention module computes attention scores using query, key, and value projections combined with rotary embeddings to effectively capture both local and long-range temporal dependencies \cite{su2024roformer}. Subsequently, the MoE feed-forward network employs a dynamic gating mechanism that routes token representations to one of four specialized temporal experts in addition to a shared expert. The MoE output for a token $h$ is computed as:

\begin{equation}
    E_{MoE}(h)=\sum^{4}_{i=1}\alpha_i\cdot E_i(h)+\alpha_s\cdot E_s(h),
\end{equation}

where the routing weights are given by $\alpha_i=softmax(W_gh)$ with $W_g\in\mathbb{R}^{4\times D}$, and shared expert gating is computed as $\alpha_s=\sigma(W_sh)$ with $W_s\in\mathbb{R}^{1\times D}$. Here, $E_i(h)$ represents the output of the $i$th specialized expert, and $E_s(h)$ is the output from the shared expert, with each expert applying an expansion via an up-projection, a non-linear SiLU activation, and a subsequent down-projection to restore the original dimensionality. Residual connections and RMS normalization are applied before the self-attention and after the MoE sub-layers to ensure stable gradient flow and robust feature integration. The final output embedding denoted $A_{embed}\in \mathbb{R}^{1\times \hat{D}}$ is then obtained by aggregating the sequence of token representations, where $\hat{D}=512$ is the embedding size. 

The encoder is pre-trained using autoregression on our statistical features sequence (from the training set only to avoid data leakage). Given the sequential input, it predicts the next five features by adding a regression head at the end of the Time-MoE encoder.

\subsubsection{Fusion and Classification}

A self-attention mechanism is employed to combine the information from both modalities effectively. First, we reshape the sum of the accelerometer features and the image features to act as the query, key, and value for the multi-head attention. Then attention embedding was computed, and this yields an attention-enhanced feature vector $F_{attn}\in \mathbb{R}^{B\times \hat{D}}$, which was then passed through multi-head classifier blocks. Each classifier head comprises multiple fully connected layers with ReLU activations, culminating in a sigmoid function to produce a probability score. 

\section{Experiments and Results}
\label{sec:experimentsandresults}

\begin{table}[t!]
    \centering
\caption{Patient Characteristics (n=126) \& Classes Distribution}
\label{tab:patient_distribution}
    \begin{tabular}{lccc}
    \hline
           &Train&Validation&  Test\\
           \hline
           \bfseries \makecell[l]{Basic information}&&& \\
           \hspace{1em}Number of patients, n&79&25&22\\
           \hspace{1em}Age, mean (SD)&56.91 (14.92)&60.8 (17.55)&63.14 (11.66)\\
           \hspace{1em}Female, n(\%)&22 (28\%)&7 (28\%)& 7(31\%)\\
           \hspace{1em}\makecell[l]{Length of stay (days),\\ median ($25^{th}$, $75^{th}$ percentile)}&\makecell[c]{13.43\\(6.33, 30.04)}&\makecell[c]{12.44\\(8.16, 29.68)}&\makecell[c]{15.00\\(7.92, 22.24)}\\
           \hspace{1em}Number of samples, n&417&133&117\\
           \hline
           \bfseries \makecell[l]{Classes (\%)}&&&\\
           \hspace{1em}Completely immobile&3 (1\%)&5 (4\%)&2 (10\%)\\
           \hspace{1em}Very limited&138 (33\%)&47 (35\%)&59 (50\%)\\
           \hspace{1em}Slightly limited&236 (57\%)&66 (50\%)&53 (45\%)\\
           \hspace{1em}No limitation&48 (12\%)&15 (11\%)&3 (3\%)\\
           \hline
    \end{tabular}
\captionsetup{font=footnotesize}
Abbreviations: n: number; SD: standard deviation
% \caption*{Abbreviations: n: number; SD: standard deviation}
\end{table}

\subsection{Experimental Setup}

\subsubsection{Dataset}

We collected our ICU dataset of 126 adult patients who agreed to participate in this research study during admission to nine specialized ICUs between 2019 and 2024. The study was approved by the University of Florida Institutional Review Board under RB201900354 and IRB202101013. Accelerometer sensor data for each participant was collected for at most 7 days, or until they were transferred or discharged from the ICU. Two types of devices were used in this study: 1) Shimmer ECG (Shimmer Sensing, Dublin, Ireland) and 2) Actigraph GTX3+ devices (ActiGraph LLC, Pensacola, FL, USA). Participants were asked to wear the sensor on their wrists and/or ankles during the data collection period. Our care provider team recorded the times when the device was removed and reapplied (for example, during surgery or bathing). In addition, mobility status data were extracted from the Electronic Health Records (EHR) and updated at each nursing shift, roughly every 12 hours.

We split the data into development and test sets by individual patient, with a ratio of 8:2, ensuring that the class distribution was preserved across all subsets. Then, we further partitioned the development set into a train and validation set for model selection and hyperparameter tuning with the same ratio. We computed and analyzed the distribution of patient characteristics and class distribution (Table \ref{tab:patient_distribution}). 

\subsubsection{Method Comparison}

We compared our model with multiple existing methods. We first benchmarked our proposed method against conventional approaches. Following the baseline setup from \cite{davoudi2019intelligent}, we used the average activity counts along each axis ($x$, $y$, and $z$) during 12-hour windows as features to predict mobility using machine learning (ML) models. Additionally, we implemented GRU (Gated Recurrent Unit), the Transformer model, and Time-MoE 200M on the accelerometer sequences to provide further performance comparisons. Next, we performed an ablation study to evaluate the performance using only the spectrogram images and the performance when using only the sequential features. We also tested the performance using the accelerometer data gathered from the ankle. Our performances were evaluated by the area under the receiver operating characteristic curve (AUROC). 

\subsubsection{Implementation details}

We first pre-trained our Time-MoE encoder in an autoregressive manner on accelerometer statistical feature sequences. This approach leverages the temporal dependencies in the data, accelerating convergence and enhancing overall performance. For the image encoder, we used the pre-trained weights of ResNet.

All experiments, including pre-training, were conducted on an NVIDIA A100-SXM4-80GB GPU, with a batch size 16 to fully utilize the available GPU memory. To prevent overfitting, we employed an early-stopping strategy with a patience of 7 epochs. The best model was selected based on the highest AUROC achieved on the validation set.

\subsection{Results}

\begin{figure}[t]
\centering
\includegraphics[width=0.8\textwidth]{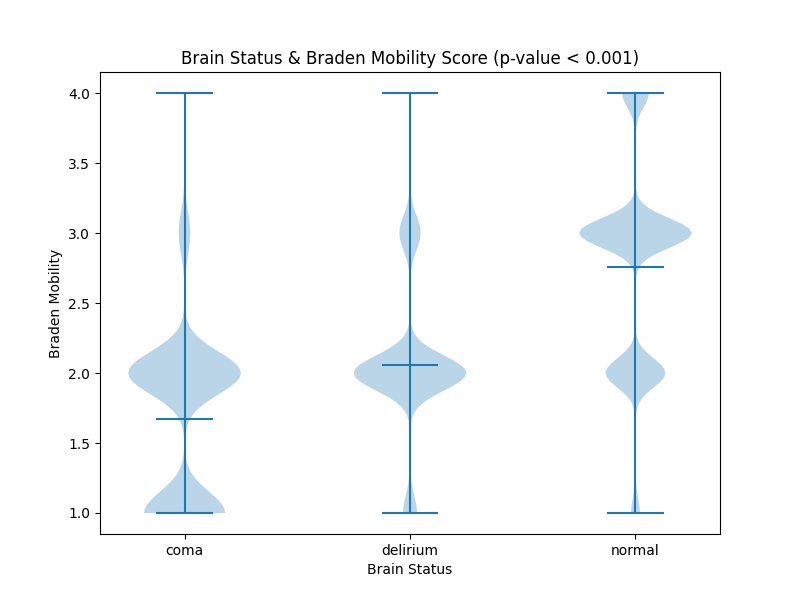}
\caption{Violin plot depicting the distribution of Braden Mobility Scores across three different brain statuses: "coma" "delirium" and "normal". } \label{fig2}
\end{figure}

\subsubsection{Classification Results}

\begin{table}[t]
\footnotesize
\centering
\caption{\textbf{Mobility} Classification Performance Comparison}
\label{tab:result_mobility}
\begin{tabular}{lccccc}
\hline
& \multicolumn{5}{c}{\bfseries AUROC (95\% CI)}\\
\bfseries Method & \makecell[c]{Completely\\Immobile} & \makecell[c]{Very\\Limited} & \makecell[c]{Slightly\\Limited} & \makecell[c]{No\\Limitation} & Overall\\
\hline
\makecell[l]{ML w/ activ.\\counts} 
  & \makecell[c]{0.65\\(0.38-0.81)} 
  & \makecell[c]{0.68\\(0.62-0.74)} 
  & \makecell[c]{0.61\\(0.56-0.68)} 
  & \makecell[c]{0.50\\(0.41-0.68)} 
  & \makecell[c]{0.61\\(0.52,0.68)}\\
\makecell[l]{GRU w/\\accel seq.}  
  & \makecell[c]{0.70\\(0.63-0.79)} 
  & \makecell[c]{0.75\\(0.67-0.83)} 
  & \makecell[c]{0.76\\(0.69-0.83)} 
  & \makecell[c]{0.35\\(0.06-0.85)} 
  & \makecell[c]{0.65\\(0.54,0.76)}\\
\makecell[l]{Transformer\\w/ accel seq.}  
  & \makecell[c]{0.52\\(0.42-0.62)} 
  & \makecell[c]{0.73\\(0.64-0.82)} 
  & \makecell[c]{0.76\\(0.69-0.84)} 
  & \makecell[c]{0.28\\(0.18-0.44)} 
  & \makecell[c]{0.54\\(0.45,0.64)}\\
\makecell[l]{Time-MoE\\w/ accel seq.} 
  & \makecell[c]{0.64\\(0.47-0.80)} 
  & \makecell[c]{0.71\\(0.60-0.81)} 
  & \makecell[c]{0.73\\(0.63-0.81)} 
  & \makecell[c]{0.81\\(0.62-0.96)} 
  & \makecell[c]{0.72\\(0.65,0.80)}\\
\hline
\makecell[l]{MELON w/o\\Spec.}
  & \makecell[c]{0.80\\(0.67-0.91)} 
  & \makecell[c]{0.76\\(0.66-0.85)} 
  & \makecell[c]{0.78\\(0.69-0.85)} 
  & \makecell[c]{0.81\\(0.64-0.94)} 
  & \makecell[c]{0.78\\(0.72,0.83)}\\
\makecell[l]{MELON w/o\\accel seq.}
  & \makecell[c]{0.23\\(0.04-0.45)} 
  & \makecell[c]{0.77\\(0.69-0.84)} 
  & \makecell[c]{0.74\\(0.67-0.83)} 
  & \makecell[c]{0.81\\(0.69-0.91)} 
  & \makecell[c]{0.63\\(0.56,0.71)}\\
\textbf{MELON}  
& \makecell[c]{\textbf{0.83}\\\textbf{(0.68-0.94)*}} 
  & \makecell[c]{\textbf{0.81}\\\textbf{(0.75-0.89)*}} 
  & \makecell[c]{\textbf{0.80}\\\textbf{(0.74-0.87)*}} 
  & \makecell[c]{\textbf{0.84}\\\textbf{(0.67-0.96)*}} 
  & \makecell[c]{\textbf{0.82}\\\textbf{(0.75,0.88)*}}\\
\hline
\end{tabular}
\captionsetup{font=footnotesize} % Adjust font size of captions globally
Abbreviations: ML: machine learning; CI: Confidence interval; accel: Accelerometer; seq.: sequence; Spec.: spectrogram.; *: p-value < 0.001 compared to baseline. P-values are based on pairwise Wilcoxon rank sum tests. 
% \caption*{Abbreviations: ML: machine learning; CI: Confidence interval; accel: Accelerometer; seq.: sequence; Spec.: spectrogram.; *: p-value < 0.001 compared to baseline. P-values are based on pairwise Wilcoxon rank sum tests. }

\end{table}

We evaluated both conventional approaches, i.e., the baseline ML model (Logistic Regression), GRU, and the transformer (Table \ref{tab:result_mobility}). Our proposed model MELON outperformed others in predicting "Completely Immobile", "Very Limited", "Slightly Limited", and "No Limitation" with AUROC (95\% C.I.) of 0.83 (0.68-0.94), 0.81 (0.75-0.89), 0.80 (0.74-0.87), and 0.84 (0.67-0.96), respectively. "Very Limited" and "Slightly Limited" were the two dominant classes, resulting in stable performance across all experimental setups. In contrast, "Completely Immobile" and "No Limitation" were rare and imbalanced categories. Notably, our proposed model demonstrated robust performance in predicting these less-represented classes. 

In our ablation study, we evaluated the model's performance by separately removing either the accelerometer sequence features or the spectrogram features. Our results showed a significant drop in performance when either modality was excluded. This further confirms that combining the global and fine-granted information enhances the model's ability.

Additionally, we compared models trained on accelerometer data collected from the patient's wrist and ankle. The results indicated that wrist-mounted data exhibited strong predictive power, effectively capturing mobility patterns. In contrast, ankle-mounted data failed to differentiate mobility differences. 

\subsubsection{Delirium Analysis}

Mobility of the patient has been approved as a potential cause of cognitive impairment such as delirium. We have tested the correlation between the four-scaled mobility assessment in our study, i.e., Braden Mobility score, and the brain status of the patients. We found a significant difference across three brain status groups (normal, delirium, and coma) with the Kruskal-Wallis p-value < 0.001. Fig. \ref{fig2} shows the distribution of the Braden Mobility scores across three groups. "Coma" shows most data points concentrated around lower mobility scores (around 1–2), suggesting limited mobility. "Delirium" displays a moderate distribution mostly around the mid-level scores. "Normal" has a distribution leaning toward higher mobility scores (around 3–4), indicating higher mobility levels.

\section{Discussion and Conclusion}
\label{sec:conclusion}

Our study demonstrates that MELON—a novel multimodal dual-branch framework combining accelerometer statistical sequences with frequency-domain spectrogram features—can effectively predict long-term mobility levels in ICU patients using wrist sensor data alone. This fusion of temporal and spectral pattern recognition holds promise for objective, continuous mobility monitoring in critical care settings.

In clinical practice, continuous monitoring of patient mobility is essential for guiding timely rehabilitation. Our work suggests that wearable wrist sensors alone can provide reliable predictive power of long-term mobility. A single-sensor solution minimizes patient discomfort, simplifies sensor management, and reduces deployment costs, thereby facilitating widespread adoption in busy critical care environments. Moreover, our analysis reveals a strong correlation between patient mobility and brain status, aligning closely with existing findings in the literature \cite{siegel2024leveraging,balas2013implementing}. This relationship highlights the potential of our methodology to be further extended into delirium detection and early alerting systems, providing an additional valuable tool for patient care.

However, the findings are constrained by the limited cohort (n=126) and class imbalance affecting rare activity detection. Additionally, while our model achieved high performance in controlled evaluations, its performance in a broader, more heterogeneous ICU scenarios remains to be validated. Future work will focus on expanding our dataset through pre-training on larger public human activity recognition datasets and integrating additional modalities, especially depth imaging \cite{siegel2024leveraging}, to enhance robustness, prioritizing computational efficiency for real-world ICU deployment.

In conclusion, MELON offers a promising approach for the continuous and objective assessment of patient mobility in the ICU. By addressing current limitations and incorporating further clinical data and modalities, our framework could significantly contribute to personalized patient management and improved clinical outcomes in critical care.

%
% ---- Bibliography ----
%
% BibTeX users should specify bibliography style 'splncs04'.
% References will then be sorted and formatted in the correct style.
%
\bibliographystyle{splncs04}
\bibliography{mybibliography}
\end{document}